%% file: main.tex
\documentclass[10pt,twocolumn,letterpaper]{article}

\usepackage[pagenumbers]{wacv}

\usepackage{graphicx}
\usepackage{amsmath}
\usepackage{amssymb}
\usepackage{booktabs}
\usepackage{enumitem}

\usepackage[pagebackref,breaklinks,colorlinks]{hyperref}

\usepackage[capitalize]{cleveref}
\crefname{section}{Sec.}{Secs.}
\Crefname{section}{Section}{Sections}
\Crefname{table}{Table}{Tables}
\crefname{table}{Tab.}{Tabs.}

\def\wacvPaperID{2500}
\def\confName{WACV}
\def\confYear{2025}

\usepackage{ulem}
\usepackage[export]{adjustbox}
\newcommand{\supplementary}{Supplementary Material}

\usepackage[accsupp]{axessibility}

\begin{document}

\title{No Annotations for Object Detection in Art through Stable Diffusion}

\author{
Patrick Ramos$^1$ \quad Nicolas Gonthier$^2$ \quad Selina Khan$^3$ \quad Yuta Nakashima$^1$ \quad Noa Garcia$^1$ \\
$^1$Osaka University \quad $^2$Univ Gustave Eiffel, ENSG, IGN, LASTIG, France \quad $^3$University of Amsterdam \\
{\tt\small \{patrickramos@is., n-yuta@, noagarcia@\}ids.osaka-u.ac.jp,} \\
{\tt\small nicolas.gonthier@ign.fr, selinajasmin@gmail.com}
}
\maketitle

\begin{abstract}
   Object detection in art is a valuable tool for the digital humanities, as it allows for faster identification of objects in artistic and historical images compared to humans. However, annotating such images poses significant challenges due to the need for specialized domain expertise. We present NADA (\underline{n}o \underline{a}nnotations for \underline{d}etection in \underline{a}rt), a pipeline that leverages diffusion models' art-related knowledge for object detection in paintings without the need for full bounding box supervision. Our method, which supports both weakly-supervised and zero-shot scenarios and does not require any fine-tuning of its pretrained components, consists of a class proposer based on large vision-language models and a class-conditioned detector based on Stable Diffusion. NADA is evaluated on two artwork datasets, ArtDL 2.0 and IconArt, outperforming prior work in weakly-supervised detection, while being the first work for zero-shot object detection in art. Code is available at
   \begin{center}
    \vspace{-10pt}
     {\footnotesize \url{https://github.com/patrick-john-ramos/nada}}     
   \end{center}
   \vspace{-10pt}
\end{abstract}

\input{sections/01-intro}
\input{sections/02-relatedwork}
\input{sections/03-method}
\input{sections/04-experiments}
\input{sections/05-conclusion}

{\small
\bibliographystyle{ieee_fullname}
\bibliography{main}
}

\newpage
\input{sections/06-appendix}

\end{document}

%% file: sections/01-intro.tex
\section{Introduction}

Performance in object detection in paintings, which has applications such as art captioning \cite{bai2021explain,cetinic2021iconographic,lu2022data}, art visual question answering \cite{garcia2020dataset}, art visual pattern discovery \cite{shen2019discovering}, musicological studies \cite{ibrahim2022few}, or art exploration \cite{meyer2024algorithmic}, lags behind traditional object detection in photographs \cite{ren2015faster,carion2020end,liu2021swin,li2022grounded}. While traditional object detection enjoys success from large-scale annotated datasets such as MS-COCO \cite{lin2014microsoft} and OpenImages \cite{li2023open}, these datasets are comprised mostly of natural images, limiting their use to other domains, \eg art images. Paintings may contain objects that might not be of interest to standard detectors and usually portray them in a different style, documented as the cross-depiction problem \cite{cai2015beyond,cai2015cross}. This domain gap can be addressed by training on datasets predominantly, if not completely, composed of non-natural images; however, annotating art images for object detection (\ie with bounding box annotations) is time-consuming and requires domain expertise. For example, in Christian iconography, an annotator must be able to distinguish between St.~Francis and St.~Dominic. While these classes can be distinguished by their associated symbols as described in online iconography databases such as Iconclass\footnote{\scriptsize\url{https://iconclass.org/}}, a deep familiarity with these relationships is still needed to annotate efficiently. As a result, existing object detection datasets in art \cite{wu2014learning,westlake2016detecting,gonthier2018weakly,milani2022proposals,reshetnikov2022deart} are much smaller than object detection datasets in natural images or only contain image-level annotations for training. 

\input{figures/out_of_dataset_samples}

In response to these limitations, various methods have been proposed to minimize the supervision required for object detection in paintings, bypassing the need for fully annotated bounding boxes around objects of interest. A first step towards reducing supervised data is to tackle the task as \textit{weakly-supervised} \cite{gonthier2018weakly,milani2022proposals}, where object detectors are trained using only image-level labels rather than detailed object bounding boxes. Additionally, reducing annotations can be taken a step further with a \textit{zero-shot} setting, where no annotations (neither bounding boxes nor class labels) are used. Due to the challenging nature of the zero-shot approach, it has not yet been explored in the art domain. 

We address this gap by introducing NADA (\underline{n}o \underline{a}nnotations for \underline{d}etection in \underline{a}rt), an application for object detection in paintings that reduces the need for supervision and detects objects in both weakly-supervised and zero-shot settings. NADA, which leverages the inherent knowledge of art in computer vision models trained on vast amounts of data, consists of two modules: a \textit{class proposer}, which, given an image of a painting and a list of potential classes, predicts the objects present in the image; and a \textit{class-conditioned detector}, which locates the objects in the painting based on the predicted classes. The class proposer can be adapted according to the desired level of supervision. If image-level classes are available, (\ie, weakly-supervised setting), a lightweight classifier is trained to classify images from their CLIP\cite{radford2021learning} embeddings. In contrast, if no annotations are available at all (\ie, zero-shot setting), the class proposer relies on a vision-language model to predict the classes present in the image. The predicted classes are used by the class-conditioned detector, which leverages the generative capabilities of diffusion models \cite{ho2020denoising,nichol2022glide,rombach2022high,ramesh2022hierarchical}, particularly Stable Diffusion\cite{rombach2022high}, to operate independently of the level of supervision. The classes are used to create an input prompt for regenerating the original image with the diffusion model. Given that diffusion models are trained on a large number of art images \cite{schuhmann2022laion} (meaning they may be familiar with objects of interest to paintings) and have been shown to contain knowledge useful for style analysis in art \cite{wu2023not} and segmenting objects in natural images \cite{yoshihashi2023attention,ma2023diffusionseg,wang2023diffusion}, we extract and segment the cross-attention maps to generate object bounding boxes, effectively detecting the objects within the painting.

NADA is quantitatively evaluated on two art object detection datasets: ArtDL~2.0 \cite{milani2022proposals} and IconArt \cite{gonthier2018weakly}. In the weakly-supervised setting, NADA outperforms prior work on ArtDL~2.0 and stays competitive with other methods on IconArt. Meanwhile, NADA presents the first results for zero-shot object detection. Our ablation study isolates the influence of the class proposer by evaluating detection when labels are already known, boosting performance on both datasets and showing that the diffusion-based class-conditioned detector localizes objects in paintings effectively, but is reliant on accurate class proposals. Lastly, we showcase the applicability of NADA by detecting uncommon objects in standard object detector datasets, such as dragons or swords, \textit{in the wild}, as shown in \cref{fig:samples}.

%% file: figures/out_of_dataset_samples.tex
\begin{figure}[tb]
  \centering
  \includegraphics[height=1.9cm, valign=c]{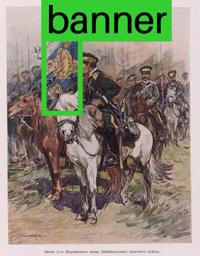}
  \includegraphics[height=1.9cm, valign=c]{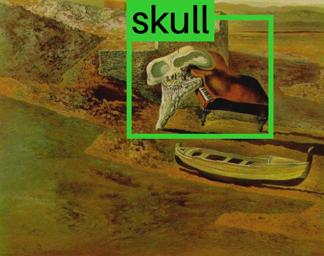}
  \includegraphics[height=1.9cm, valign=c]{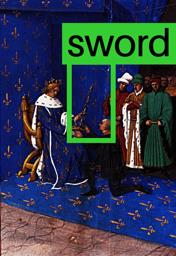}
  \includegraphics[height=1.9cm, valign=c]{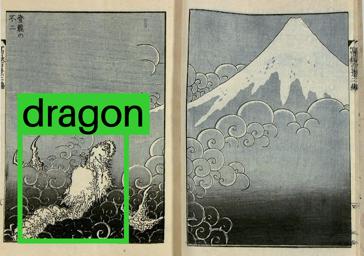}
  \vfill
  \includegraphics[height=1.9cm, valign=c]{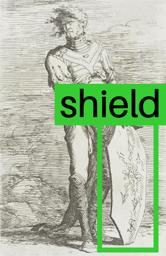}
  \includegraphics[height=1.9cm, valign=c]{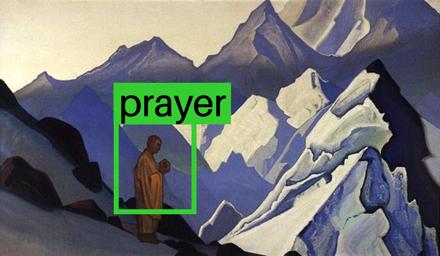}
  \includegraphics[height=1.9cm, valign=c]{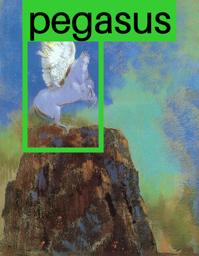}
  \includegraphics[height=1.9cm, valign=c]{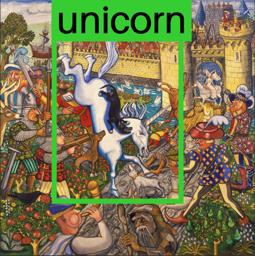}
  \caption{Art object detection \textit{in the wild} with NADA's class-conditioned detector.}
  \label{fig:samples}
\end{figure}

%% file: sections/02-relatedwork.tex
\section{Related work}

\paragraph{Object detection in art}

Localizing and recognizing objects in art presents some unique challenges compared to object detection in natural images \cite{redmon2016you,ren2015faster}, primarily due to the interest in objects that are not common in natural images and the cross-depiction problem. While differences in style between natural images and paintings can contribute to the difficulty of object detection in art, previous work \cite{crowley2016art,ahmad2023toward,ginosar2015detecting,westlake2016detecting,inoue2018cross} have shown that transfer learning and domain adaptation techniques can perform reasonably well in bridging this gap. A survey on this topic is available in \cite{bengamra2024comprehensive}. However, an important challenge arises when the classes to be detected, such as mythological creatures like \textit{dragons} and \textit{angels} or historical figures like \textit{Napoleon Bonaparte}, are entirely different from those in natural image datasets, making transfer learning and domain adaptation techniques less effective. This problem is exacerbated by the cost of annotating bounding boxes for such novel classes. One approach is to leverage descriptions to address the knowledge gap \cite{khan2024context}, however this still requires painting descriptions. To address the difficulty of annotating data, other methods approach art object detection as a weakly supervised task.

In this setting, weakly supervised detectors are trained using only image labels, which indicate the classes present in the image but not their locations. This approach has been extensively studied for natural images, with methods based on end-to-end training of modified object detectors   \cite{bilen2016weakly,wan2019c,ren2020instance,seo2022object}. In the domain of art, Gonthier \etal~\cite{gonthier2018weakly} treated the task as a multiple-instance-learning (MIL) problem by training a classifier on top of Faster R-CNN \cite{ren2015faster} bounding box features and objectness scores. This method was extended in \cite{gonthier2022multiple} with a multi-layer model to boost the performance at minimal extra cost. Milani \etal~\cite{milani2022proposals} used pseudo-data by creating bounding boxes from class activation maps (CAMs) extracted from a ResNet-50 \cite{he2016deep} fine-tuned on the target domain. At the extreme, \cite{madhu2022one} proposed a one-shot learning method using a modified co-attention and co-excitation framework \cite{hsieh2019one} and data contextualization. Our approach, NADA, extends prior work which only goes as far as weak-supervision by also proposing a zero-shot method that does not require training on a target dataset.

\vspace{-12pt}
\paragraph{Locating objects with diffusion models}
Diffusion models\cite{ho2020denoising} are image generation models consisting of denoising auto-encoders that are often conditioned on text inputs\cite{nichol2022glide,rombach2022high,ramesh2022hierarchical}. Despite their main purpose being image generation, diffusion models have been leveraged for image segmentation \cite{karazija2023diffusion,yoshihashi2023attention,wu2023diffumask,ma2023diffusionseg,wang2023diffusion} and object detection \cite{feng2024instagen} in two main ways. The first way consists of generating synthetic training images, extracting attention maps from the diffusion model during generation, and converting these attention maps into pseudo-segmentation masks \cite{yoshihashi2023attention,wu2023diffumask,ma2023diffusionseg,wang2023diffusion}. The second approach attaches detection or segmentation modules directly to the internal representations of diffusion models \cite{baranchuk2022labelefficient,ma2023diffusionseg,kondapaneni2024text,li2023open,feng2024instagen} by obtaining noise corresponding to the input image through noising or diffusion inversion \cite{song2020denoising}, denoising the noise, and extracting the intermediate representations to predict bounding boxes or segmentation masks using an encoder \cite{baranchuk2022labelefficient} or decoder head \cite{ma2023diffusionseg,kondapaneni2024text}, sometimes combined with text features \cite{li2023open,feng2024instagen}.

\begin{figure*}[tb]
  \centering
  \includegraphics[width=0.95\linewidth]{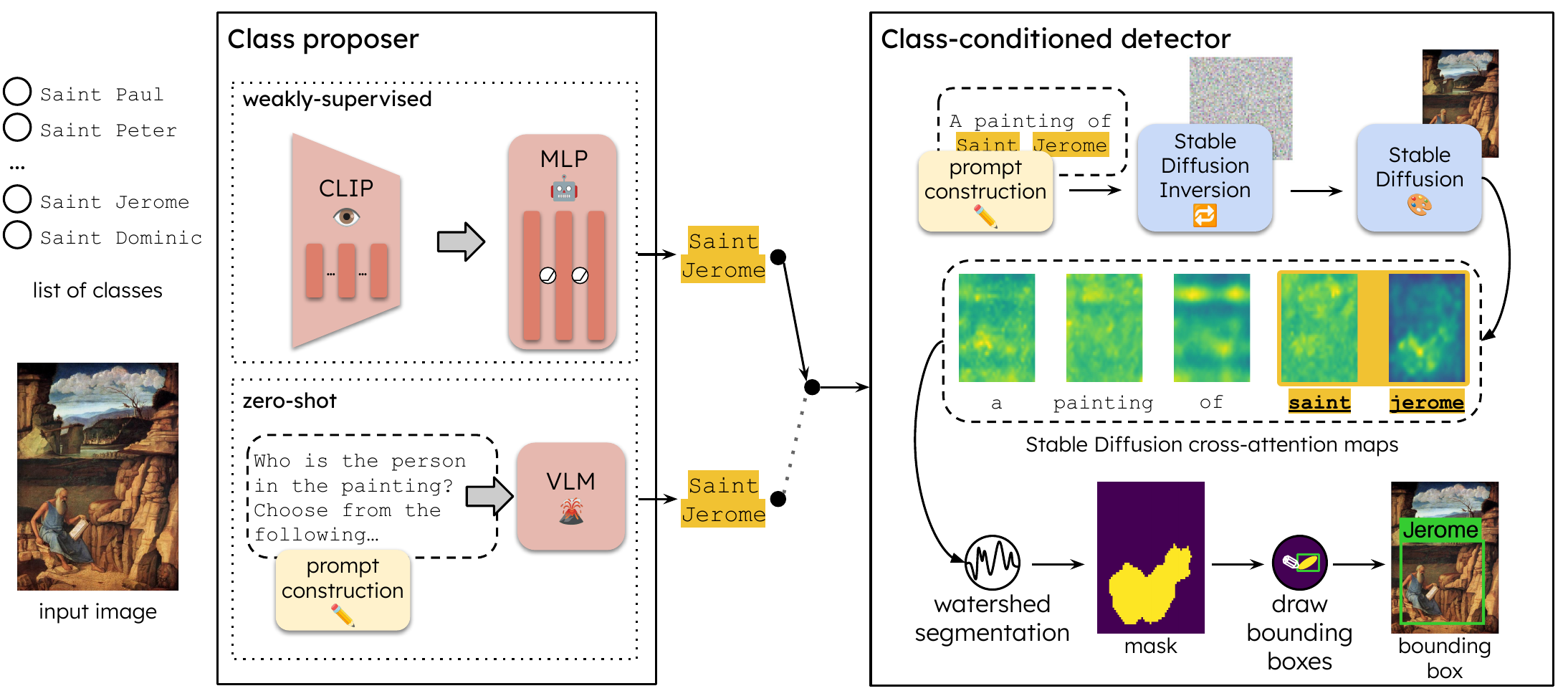}
  \caption{NADA consists of predicting classes from a painting with a class proposer and extracting bounding boxes for the predicted classes with a class-conditioned detector. The class proposer can operate in a weakly-supervised or a zero-shot setting. The class-conditioned detector leverages Stable Diffusion to extract bounding boxes by inverting and regenerating the painting conditioned on an input prompt. The cross-attention maps from the predicted class are aggregated and processed with watershed segmentation to find the bounding box.}
  \label{fig:overview}
\end{figure*}

Of prior work, DiffusionSeg \cite{ma2023diffusionseg} is the most similar to our approach as it reports results on extracting segmentation masks from attention maps obtained from real images using diffusion inversion and without training on synthetic data. NADA differs from that study as we focus instead on object detection, propose a simpler method of extracting bounding boxes, and specifically explore the suitability of leveraging Stable Diffusion's knowledge for art images.

%% file: sections/03-method.tex
\section{Method}
\label{sec:method}

\Cref{fig:overview} provides an overview of our method. Given an image $i$ and a set $\mathcal{L} = \{l\}$ of classes $l$ of possible objects\footnote{We use $l$ to denote both the class label (\eg, \texttt{mary}) and its textual representation (\eg, ``Mary'') interchangeably depending on the context.}, NADA predicts a set $\mathcal{B}$ of bounding boxes in $i$ containing objects in $\mathcal{L}$. Following previous work~\cite{milani2022proposals}, NADA is divided into a \textit{class proposer} to predict a plausible set $\mathcal{L}' \subseteq \mathcal{L}$ of classes from $i$, and a \textit{class-conditioned detector} to predict $\mathcal{B}$ from $i$ and $\mathcal{L}'$. The class-conditioned detector leverages the art-related knowledge in Stable Diffusion to localize a given object in the image. Our class proposer module identifies $\mathcal{L}'$ without training with bounding box annotations, \ie weakly-supervised or zero-shot.

\subsection{Class proposer}

We formulate the task of finding the set $\mathcal{L}'$ of class proposals that are likely to appear in $i$ as a classification task. In the weakly-supervised version of our pipeline, we use class labels to train a simple classifier to predict which classes are in $i$. In the zero-shot version of our method, we task a frozen vision language model (VLM) to predict classes without any training.

\vspace{-12pt}
\paragraph{Weakly-supervised class proposal (WSCP)} 
We use a frozen CLIP image encoder followed by a multi-layer perceptron (MLP) to classify $i$. We formulate this as $\mathcal{L}' = \text{MLP}(\text{CLIP}(i))$. We train the MLP either with a single-label classification task using cross-entropy loss or with a multi-label classification task using binary cross-entropy loss. We leverage domain knowledge of the target art datasets to choose which task and loss to train the MLP with.

\vspace{-12pt}
\paragraph{Zero-shot class proposal (ZSCP)}
We use a frozen VLM to classify $i$ without any training. Given $\mathcal{L}$, we design a prompt $q$ to ask the VLM to identify all $l \in \mathcal{L}$ in $i$. Class proposals are given by $\mathcal{L}' = P(\text{VLM}(i,q))$, where $P$ is a simple text post-processing function. 

\subsection{Class-conditioned detector}

This module takes $i$ and each class label $l \in \mathcal{L}'$ to predict a set $\mathcal{B}_l$ of bounding boxes that contain an object of class $l$. To leverage art knowledge in Stable Diffusion, we obtain cross-attention maps from an input image $i$ by performing a diffusion process (\ie, Stable Diffusion inversion) followed by a reverse diffusion process (\ie, Stable Diffusion), both guided by a prompt $p$ containing the class label $l$. The reverse diffusion process provides a cross-attention map between each token in $p$ and each patch in $i$, which identifies which patches in $i$ are associated with $l$. Letting $A_l$ denote the cross-attention map for $l$, we apply watershed segmentation \cite{soille1990automated,neubert2014compact} to $A_l$ to find regions relevant to $l$. Then, bounding boxes are computed from each of the regions. Specific details for each of these processes are provided below.

\vspace{-12pt}
\paragraph{Prompt construction}
Given a label $l \in \mathcal{L}'$, we construct a prompt $p$ that describes the image and contains $l$. During prompt construction, we modify labels to make them more concrete \eg concretizing the label \textit{nudity}, a state of existence, to the more perceivable \textit{naked person}. We also generalize some labels depending on the scope of the domain, such as generalizing \textit{Child Jesus} to the simpler concept \textit{baby} if it is the only baby among the objects of interest. Note that label generalization is one area of improvement as there are cases where it may confuse classes \eg generalizing \textit{Child Jesus} to \textit{baby} when \textit{Child St. John the Baptist} is also in the data.

\vspace{-12pt}
\paragraph{Stable Diffusion inversion}
Our method relies on the cross-attention maps of Stable Diffusion $D$ as it produces the input image $i$ with the prompt $p$. However, having a model designed to generate synthetic images output existing ones is less straightforward. To allow $D$ to produce $i$, we first invert the image using null-text inversion \cite{mokady2023null}, which generates noise $n$ from an image-prompt pair that reproduces $i$ when $p$ is fed to $D$. We denote the inversion process as $n = N_D(i,p)$, where $N_D$ is the inversion function. Note that the null-text inversion process is conducted over several steps, making it time-consuming and another area of improvement.

\vspace{-12pt}
\paragraph{Stable Diffusion reconstruction}
With the noise obtained from inversion, we use Stable Diffusion to generate $i$. The reverse diffusion process is denoted as $i, \{A'_{jk}\}_{jk} = D(n, p)$, which produces the cross attention map $A'_{jk}$ between $p$ and $i$ from the $k$-th cross-attention block at time step $j$ of the reverse diffusion process ($k = 1,\dots,K$, $j=1,\dots,J$). We discard the reconstructed $i$ and keep only the attention maps $\{A'_{jk}\}_{jk}$.

\vspace{-12pt}
\paragraph{Extracting image-text cross-attention maps}
The cross-attention map $A'_k \in \mathbb{R}^{T \times H \times W}$ encompasses the attention weights between each token in $p$ and each patch in $i$, where $T$ refers to the number of tokens in $p$, and $H$ and $W$ are the numbers of patches comprising the height and width of the attention map. $A_t \in \mathbb{R}^{H \times W}$ is the average of the cross-attention maps across all layers and time steps of the network corresponding to token $t \in p$, given by:
\begin{align}
    A_t = \frac{1}{JK}\sum_{j,k} A'_{jkt},
\end{align}
where the summation is computed over $k = 1,\dots,K$ and $j=1,\dots,J$, and $A'_{jkt} \in \mathbb{R}^{H \times W}$ is the map that contains the relevance of $t$ to the image patches. As $l$ may consist of multiple tokens (\eg, $l = \texttt{``john the baptist''}$ may consist of \texttt{john}, \texttt{the}, and \texttt{baptist}), we again average all maps associated with $l$ to obtain the attention map $A_l$ for $l$, \ie,
\begin{align}
    A_l = C(\frac{1}{|l|}\sum_{t \in l} A_t),
\end{align}
where $|l|$ denotes the number of tokens in $l$ and $C$ is a clamp function that clamps the map's values between 0 and 1.

\vspace{-12pt}
\paragraph{Extracting bounding boxes}

\begin{figure}[tb]
  \centering
  \includegraphics[width=0.95\linewidth]{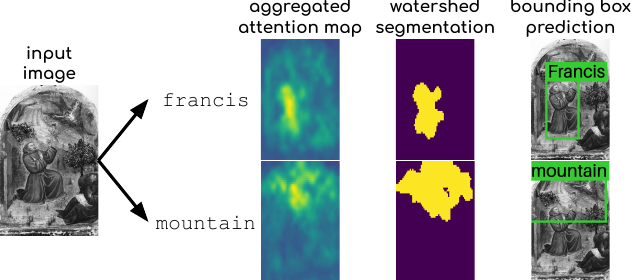}
  \caption{Bounding box extraction from attention maps.}
  \label{fig:box_extraction}
\end{figure}

As demonstrated by prior work \cite{tang2023daam}, the attention map $A_l$ gives a larger value to regions relevant to $l$. Based on this, we can extract regions by thresholding it. We use Otsu's method \cite{otsu1975threshold} to binarize the attention map and segment any region of interest with watershed segmentation \cite{neubert2014compact,soille1990automated}. We identify the boxes enclosing the masked regions and take these as the predicted bounding boxes $\mathcal{B}_l$ and group them as $\mathcal{B} = \bigcup_l \mathcal{B}_l$. An example of the process of extracting bounding boxes from an input painting is shown in \cref{fig:box_extraction}.

%% file: sections/04-experiments.tex
\section{Experiments}
\label{sec:experiments}

\paragraph{Evaluation datasets}
We evaluate NADA on two standard object detection datasets in art: ArtDL~2.0 \cite{milani2022proposals} and IconArt \cite{gonthier2018weakly}. ArtDL~2.0 contains ten classes of Christian icons taken from the Iconclass database. The dataset comprises $21,673$ images annotated with classes for training, along with $2,632$ test images annotated with class labels only and $808$ test images annotated with bounding boxes. Furthermore, $2,628$ labeled images and $1,625$ bounding-box annotated images are provided as validation sets. Similarly, IconArt focuses on seven classes of Christian iconography, with $1,421$ images for classification training, $2,053$ images for classification evaluation, $610$ images for classification validation, and $1,480$ images for detection evaluation. A summary of the evaluation datasets is provided in the \supplementary.

\input{tables/weakly_supervised_classification}

\vspace{-12pt}
\paragraph{Implementation details}
In the weakly-supervised setting, we use a CLIP ViT-B/32\footnote{\scriptsize\url{https://huggingface.co/openai/clip-vit-base-patch32}} as the CLIP image encoder. We use two layers for the MLP for ArtDL~2.0 and three layers for IconArt. Both MLPs use a hidden size of $384$ and ReLU activation. As most images in ArtDL~2.0 contain a single object, we train with single-label classification. Meanwhile, IconArt tends to have multiple classes in each image, so we use multi-label classification for it. All MLPs are trained for $100$ epochs with AdamW \cite{loshchilov2017decoupled} optimizer and a batch size of $512$. Training hyperparameters per dataset are provided in the \supplementary.

In the zero-shot setting, we use LLaVA-NeXT-34B\footnote{\scriptsize\url{https://huggingface.co/liuhaotian/llava-v1.6-34b}} \cite{liu2024visual} as VLM with two types of prompts as input:

\begin{itemize}[noitemsep,topsep=0pt]
    \item \textbf{Choice}: We query the VLM to select which classes among $\mathcal{L}$ are present in the image. The text post-processing function $P$ simply consists of extracting the predicted classes from the VLM text output.

    \item \textbf{Score}: We query the VLM to provide a confidence score $s_l \in [0, 1]$ for each $l \in \mathcal{L}$, with each score indicating the likelihood that a label $l$ appears in the image. The text post-processing function $P$ consists of extracting the labels and scores and thresholding the scores with predefined $\tau$ to identify a set $\mathcal{L}' = \{l \in \mathcal{L}|s_l > \tau\}$. We tune $\tau$ on the validation split of IconArt and set $\tau=0.5$.
\end{itemize}

For the class-conditioned detector, we use Stable Diffusion 2\footnote{\scriptsize\url{https://huggingface.co/stabilityai/stable-diffusion-2-base}} for inversion and reconstruction. We perform null-text inversion\footnote{\scriptsize\url{https://github.com/google/prompt-to-prompt/}} \cite{mokady2023null} over $500$ steps and reconstruct images for $50$ steps. We consider two prompt construction methods for inversion and reconstruction:

\begin{itemize}[noitemsep,topsep=0pt]
    \item \textbf{Template}: We insert the class name into a pre-defined prompt template.

    \item \textbf{Caption}: We use the same VLM to describe the image with a caption that contains the class name. Captions that do not contain the class or contain the class at a position beyond the maximum input length of the diffusion model are prepended with a prompt template formatted with the class name.
\end{itemize}

For ArtDL~2.0, we use the Wikipedia\footnote{\scriptsize \url{https://www.wikipedia.org/}} article titles corresponding to each class as labels. For IconArt, we change some of the class names as follows:  \textit{Saint Sebastien} to \textit{person}, \textit{child Jesus} to \textit{baby}, and \textit{nudity} to \textit{naked person}. All the prompts can be found in the \supplementary.

\subsection{Weakly-supervised evaluation}

\paragraph{Baselines} 
 We compare our method against previous work on weakly-supervised object detection:

\begin{itemize}[noitemsep,topsep=0pt]
    \item[$\bullet$] PCL \cite{tang2018pcl}: It uses an MIL network on top of projected CNN features to generate proposal scores and clusters, which are used to start an iterative refinement of an instance classifier. At each refinement step, the current instance classifier is guided by proposal clusters generated from the previous step.
    \item[$\bullet$] CASD \cite{huang2020comprehensive}: It also has an MIL head and an iteratively refined instance classifier over image features, but features across input transformations and layers are aggregated to create comprehensive attention maps and are used to guide self-distillation of the detector.
    \item[$\bullet$] UWSOD \cite{shen2020uwsod}: Object locations are proposed with an anchor-based self-supervised object proposal generator. Both detection scores and boxes are progressively improved via a step-wise bounding-box fine-tuning process. A multi-rate resampling pyramid is used to combine multi-scale contextual information.
    \item[$\bullet$] CAM+PaS \cite{milani2022proposals}: A ResNet-50 is fine-tuned on the target dataset and used to extract class-activation maps (CAMs) from images. Percentiles of the CAM values are used to threshold the CAMs and bounding boxes are drawn around the salient area.
    \item[$\bullet$] Milani \cite{milani2022proposals}: CAM+PaS is used to create pseudo-ground-truth bounding boxes for a set of images and a Faster R-CNN is trained on them.
    \item[$\bullet$] MI-Max-HL \cite{gonthier2022multiple}: A pretrained Faster R-CNN is used to extract proposal embeddings and objectness scores. Embeddings are processed by a fully connected layer and a MIL classifier before being multiplied by objectness scores. The highest-scoring proposals from the MIL classifier are taken as positive predictions during weakly supervised fine-tuning.
\end{itemize}

We do not re-implement the above baselines; instead, we report results as presented in previous works \cite{milani2022proposals,gonthier2022multiple,tang2018pcl}.

\vspace{-12pt}
\paragraph{Classification results}
\label{sec:ws_classificaiton}
To evaluate classification accuracy, we report precision (P), recall (R), F1 score (F1), and classification average precision (AP) for each dataset in \cref{tab:weaksupervised_classification}. Using only a simple MLP for training,  our weakly-supervised class proposer (WSCP) achieves the highest P, F1 score and AP on IconArt. Moreover, it shows competitive performance compared to a more complex fully fine-tuned ResNet-50 (Milani) on the ArtDL~2.0 dataset, where it also obtains the best precision. In summary, our WSCP not only outperforms state-of-the-art models in terms of simplicity but also obtains competitive results, showcasing its effectiveness in weakly supervised object detection 

\input{tables/weakly_supervised_detection}

\vspace{-14pt}
\paragraph{Object detection results}
Detection results are reported in \cref{tab:weaksupervised_detection} as AP$_{50}$, which measures the area under the precision-recall curve for detections above a 0.5 intersection over union (IoU) threshold. For NADA, we report the result of the best prompt construction method per dataset. Results show that NADA achieves the highest performance on ArtDL~2.0 with an AP$_{50}$ of $45.8$. Meanwhile, on IconArt, NADA stays competitive with Milani and MI-Max-HL methods with only $0.7$ and $2.8$ AP$_{50}$ points difference, respectively, while outperforming the remaining baselines by a higher score (7.6 AP$_{50}$ points higher than the next best method). NADA achieves this while being one of only two evaluated methods that do not require training the detector. Note that we report more results in \cref{tab:weaksupervised_detection} than in \cref{tab:weaksupervised_classification} as some methods only reported detection and not classification scores.

Interestingly, superior AP classification accuracy does not necessarily translate to a better AP$_{50}$ in object detection, as previously noted in \cite{gonthier2022multiple}. This discrepancy suggests that while our Stable Diffusion-based method detects and localizes depicted classes more accurately, Faster R-CNN and its variants leverage stronger features than the off-the-shelf internal representations of Stable Diffusion. To investigate this, in \cref{sec:upperbound}, we isolate the influence of the class proposer and report the results of the class-conditioned detector when a perfect class proposal module is assumed.

\subsection{Zero-shot evaluation}
\label{sec:zeroshot}

\paragraph{Baselines} As there is no prior research on zero-shot object detection in art, and DiffusionSeg \cite{ma2023diffusionseg}, which is the most closely related work leveraging Stable Diffusion for object segmentation, does not have publicly available code for reproduction, we compare our zero-shot NADA against two baseline class proposals methods: CLIP-based and InstructBLIP-based.
\begin{itemize}[noitemsep,topsep=0pt]
    \item[$\bullet$] CLIP: We use the standard zero-shot protocol in CLIP \cite{radford2021learning}, where each test image is embedded with a pretrained CLIP image encoder and matched against a text embedded with a pretrained CLIP text encoder with the prompt \texttt{``A painting of [CLASS]''}. Any \texttt{[CLASS]} with a cosine similarity greater than $0.28$ is taken as a predicted class. This threshold is based on the CLIP filtering process of LAION-5B\cite{schuhmann2022laion}.
    
    \item[$\bullet$] InstructBLIP: We replace LLaVA with InstructBLIP-Vicuna-7B\footnote{\scriptsize\url{https://huggingface.co/Salesforce/instructblip-vicuna-7b}} \cite{dai2023instructblip}. We prompt InstructBLIP with each class individually using the query \texttt{``Is [CLASS] in the painting?''}. An output containing \texttt{``yes''} is taken as a positive prediction and any other response is considered a negative prediction. We use one prompt per class as using our choice or score prompts tended to produce irrelevant results.
\end{itemize}

Note that whereas CLIP and InstructBLIP class proposals need to use a prompt for each class and image, our zero-shot class proposer (ZSCP) uses only one prompt per image. We use the same class-conditioned detector based on Stable Diffusion for baseline detection results.

\input{tables/zero_shot_classification}

\vspace{-12pt}
\paragraph{Classification results}
Zero-shot classification results are shown in \cref{tab:zeroshot_classification}. We report the results of our ZSCP using the two types of prompts: choice and score. Results show that both choice and score prompts are not only the most efficient methods, requiring only one prompt per image compared to CLIP and InstructBLIP, which need a prompt per class and image, but also achieve the highest precision, F1 score and AP on both datasets. InstructBLIP provides the best recall, however, it also has the lowest precision on both datasets, indicating it is overpredicting classes. When comparing score to choice prompting methods, we observe inconsistent results. While the score prompt leads to more accurate predictions in terms of precision and AP for IconArt, the choice prompt leads to the best classification performance for ArtDL~2.0.

\input{tables/zero_shot_detection}

\vspace{-12pt}
\paragraph{Object detection results}
Zero-shot detection results are reported in \cref{tab:zeroshot_detection}. For NADA, we report the best result among VLM prompts and prompt construction for each dataset. On ArtDL~2.0, following the classification results, the large gap between the ZSCP and the two baselines on F1 score and AP leads to NADA obtaining the highest detection performance with an AP$_{50}$ of $21.8$, surpassing InstructBLIP by $3.2$ AP$_{50}$ points. On the IconArt dataset, despite the classification results among all the models being closer in terms of F1 score and AP, NADA is able to achieve the highest performance with an AP$_{50}$ of $15.8$, which is $7.9$ AP$_{50}$ points above InstructBLIP. Notably, the zero-shot version of NADA even outperforms some weakly-supervised methods in \cref{tab:weaksupervised_detection} while requiring no annotations whatsoever. Although its performance may lag behind state-of-the-art fully-supervised methods, it does not require any training on the target dataset.

\input{tables/prompt_construction}
\input{tables/prompt_ablation}

\subsubsection{Qualitative analysis}
\input{figures/attention_box_samples}

We show attention map visualizations and bounding boxes predicted by NADA ({\footnotesize with ZSCP}) in \cref{fig:attention_box_samples}. Stable Diffusion's knowledge of art images can indeed localize objects in paintings, as the attention maps highlight the sought labels. The bounding boxes contain the salient regions of the attention map, showing that NADA can transform attention maps into meaningful bounding boxes. Even when there are multiple subjects, NADA is capable of detecting Mary among five people (top row, second from right). One can also see that Stable Diffusion knows about the iconographic attributes of some characters, such as the arrows of Saint Sebastian\footnote{Saint Sebastian's association with arrows is a common representation in iconography and is discussed in \scriptsize \url{https://en.wikipedia.org/wiki/Saint_Sebastian}.} (top row, second from left). NADA may fail when the class is incorrect, such as misclassifying Paul as Jerome (bottom row, leftmost), however it is still able to localize the subject. NADA may also correctly identify objects but incorrectly localize them, such as identifying the wrong person as an angel (bottom row, second from right).

\subsection{Analysis and ablation studies}

\paragraph{Upper bound object detection performance}
\label{sec:upperbound}
We measure the upper-bound detection performance of NADA when assuming a perfect class proposal module that always predicts the correct classes. This NADA configuration, referred to as \emph{Oracle} allows us to discern the accuracy contribution of the class-conditioned detector and the adequacy of Stable Diffusion's cross-attention maps for art object detection. Given the correct labels, the Oracle substantially improves performance from $21.8$ (zero-shot) and $45.8$ (weakly-supervised) to $61.3$ AP$_{50}$ on the ArtDL~2.0 dataset. On the IconArt dataset, the Oracle boosts object detection results from $15.1$ (zero-shot) and $13.8$ (weakly-supervised) to $18.7$ AP$_{50}$. This implies that a large part of the object detection performance is dependent on the accuracy of the class proposer. Within the same method, better class predictions lead to better object detection performance.

\vspace{-12pt}
\paragraph{Impact of different thresholds}

\begin{figure}
    \centering
    \includegraphics[width=0.95\linewidth]{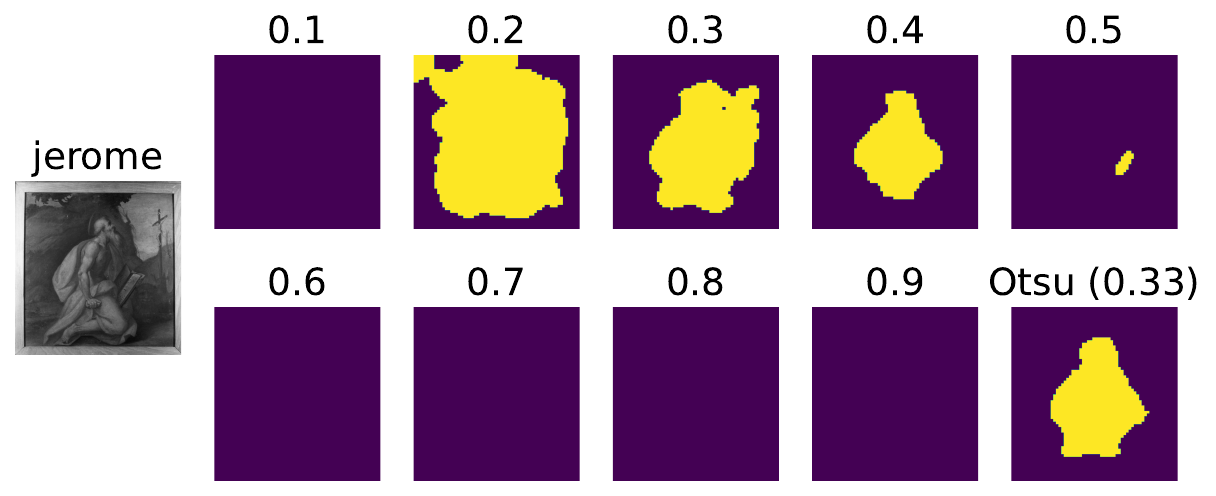}
    \caption{Mask prior to bounding box drawing for different thresholds, including Otsu's method.}
    \label{fig:otsu_examples}
\end{figure}

\begin{figure}
    \centering
    \includegraphics[width=\linewidth]{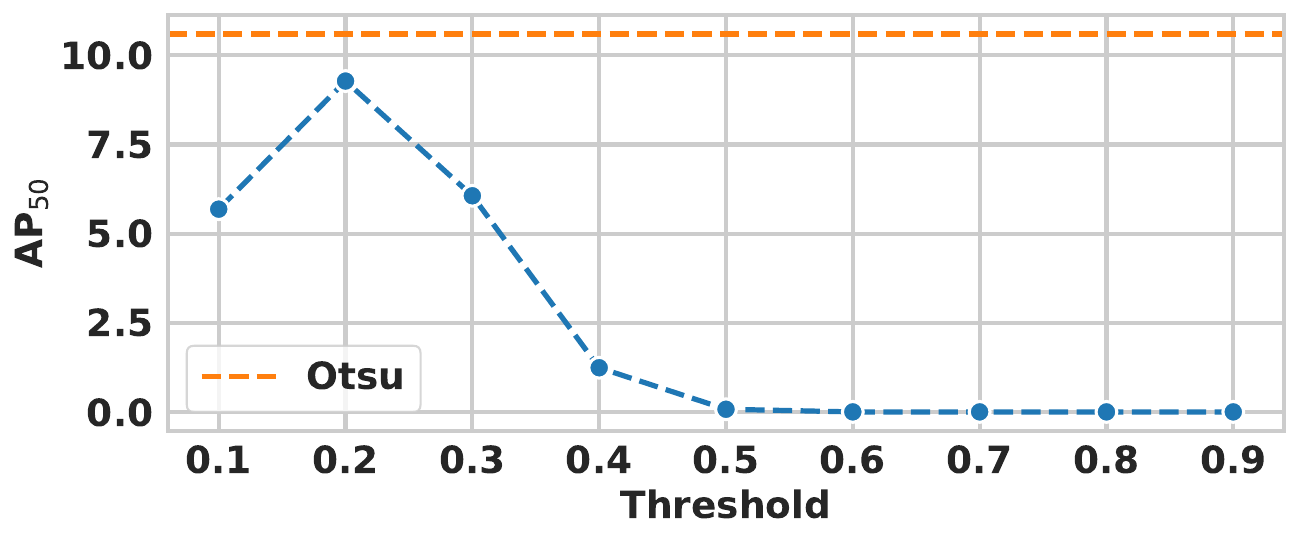}
    \caption{
    AP$_{50}$ results on the ArtDL validation detection set for varying thresholds and when using Otsu's method.}
    \label{fig:otsu_graph}
\end{figure}

We analyze the use of Otsu threshold in \cref{fig:otsu_examples,fig:otsu_graph}. \Cref{fig:otsu_examples} shows the watershed segmentation mask prior to bounding box construction for thresholds ranging from $0.1$ to $0.9$ in intervals of $0.1$. Lower thresholds lead to masks that are too large, while higher thresholds result in masks that are too small, with thresholds above $0.5$ resulting in no mask being segmented. In this example, Otsu's method determined the optimal threshold value to be $0.33$, which was not among any of the manually tested thresholds. \Cref{fig:otsu_graph} shows AP$_{50}$ performance on the validation detection split of ArtDL~2.0 for each of the thresholds. We find that the best-performing threshold is still the one determined via the Otsu's method.

\vspace{-12pt}
\paragraph{Impact of prompting method}
\Cref{tab:prompt_construction} shows the impact of NADA's prompting method on object detection performance. We observe different behaviors between the two datasets, but consistent behaviors within. Caption prompts show consistently lower performance than template prompts in ArtDL~2.0, but consistently outperform template prompts in IconArt. We initially believed this was due to ArtDL~2.0 and IconArt tending to have one and multiple classes per image respectively. However, upon checking how NADA {\footnotesize (with ZSCP)} performs on single and multi-object subsets of ArtDL~2.0 and IconArt in \cref{tab:prompt_ablation}, we find that this is not the case. Regardless of the number of classes in an image, template prompt construction improves ArtDL~2.0 detection while caption prompt construction boosts IconArt detection. It should be noted however that templates outperform captions more on ArtDL~2.0 when there is only one label ($+1.7$ AP$_{50}$) than when there are multiple ($+0.01$ AP$_{50}$) while captions outperform templates more on IconArt when there are multiple labels ($+2.9$ AP$_{50}$) than where there is only one ($+0.4$ AP$_{50}$).

\vspace{-12pt}
\paragraph{Object detection in the wild} We use NADA's class-conditioned detector with caption prompt construction to detect objects in WikiArt images \emph{in the wild}. Examples are shown in \cref{fig:samples}. These images contain subjects that are not typically considered in natural image object detectors and are even not among the classes in either ArtDL~2.0 or IconArt, while also covering a variety of styles including Renaissance, ukiyo-e, and surrealism. NADA is able to detect uncommon objects often portrayed in art such as banners and shields alongside mythological creatures such as dragons, Pegasus, and unicorns. NADA is also able to understand artistic interpretations of these classes such as a surrealist rendering of melting skull. These results indicate that NADA is capable of bridging the domain gap in both classes and styles presented by art images. Furthermore, contrary to other methods performing object detection from text inputs, our approach does not rely on Google Images search of the objects of interest \cite{crowley2015search}.

%% file: tables/weakly_supervised_classification.tex
\begin{table*}[t]
  \footnotesize
  \caption{Weakly-supervised classification results. All metrics are macro-averaged. Params indicates the number of trainable parameters.}
  \label{tab:weaksupervised_classification}
  \centering
  \setlength{\tabcolsep}{4pt}
  \begin{tabular}{@{}lrcccccccccc@{}}
    \toprule
    & & & \multicolumn{4}{c}{ArtDL 2.0 \cite{milani2022proposals}} & & \multicolumn{4}{c}{IconArt \cite{gonthier2018weakly}} \\
    \cline{4-7}
    \cline{9-12}
    Method & Params & &  P & R & F1 & AP & &  P & R & F1 & AP \\
    \midrule
    Milani \cite{milani2022proposals} & 23.4M & & 72.7 & 69.8 & \textbf{69.1} & \textbf{71.6} & & 71.7 & 61.9 & 65.6 & 73.1 \\
    MI-Max-HL \cite{gonthier2022multiple} & 3.7M & & 4.0 & \textbf{85.0} & 9.0 & 17.6 & & 24.0 & \textbf{97.0} & 36.0 & 54.0 \\
    WSCP (ours) & 0.4M & & \textbf{78.1} & 49.3 & 57.8 & 57.5 & & \textbf{80.5} & 69.6 & \textbf{74.1} & \textbf{80.7} \\
    \bottomrule
  \end{tabular}
\end{table*}

%% file: tables/weakly_supervised_detection.tex
\begin{table}[t]
  \footnotesize
  \caption{Weakly-supervised object detection results as AP$_{50}$.}
  \label{tab:weaksupervised_detection}
  \centering
    \setlength{\tabcolsep}{4pt}
  \begin{tabular}{@{}lccc@{}}
    \toprule
    Method & Train detector? & ArtDL 2.0 \cite{milani2022proposals} & IconArt \cite{gonthier2018weakly} \\
    \midrule
    PCL \cite{tang2018pcl} & $\checkmark$ & 24.8 & 5.9 \\
    CASD \cite{huang2020comprehensive} & $\checkmark$ & 13.5 & 4.5 \\
    UWSOD \cite{shen2020uwsod} & $\checkmark$ & 7.6 & 6.2 \\
    CAM+PaS \cite{milani2022proposals} & $\checkmark$ & 40.3 & 3.2 \\
    Milani \cite{milani2022proposals} & $\checkmark$ & 41.5 & \textbf{16.6} \\
    MI-Max-HL \cite{gonthier2022multiple} & $\times$ & 8.2 & 14.5 \\
    NADA {\footnotesize(with WSCP)} & $\times$ & \textbf{45.8} & 13.8 \\
    \bottomrule
  \end{tabular}
\end{table}

%% file: tables/zero_shot_classification.tex
\begin{table*}[t]
  \footnotesize
  \caption{Zero-shot classification results. Num. prompts indicate the number of prompts per image. All  metrics are macro-averaged.}
  \label{tab:zeroshot_classification}
  \centering
    \setlength{\tabcolsep}{4pt}
  \begin{tabular}{@{}lrcccccccccc@{}}
    \toprule
    & & & \multicolumn{4}{c}{ArtDL 2.0} & & \multicolumn{4}{c}{IconArt} \\
    \cline{4-7}
    \cline{9-12}
    Class proposal & Num. prompts & &  P & R & F1 & AP & &  P & R & F1 & AP \\
    \midrule
    CLIP & Num. classes & & 30.2 & 27.7 & 15.2 & 14.6 & & 65.8 & 55.1 & 49.9 & 48.5 \\
    InstructBLIP  & Num. classes & & 27.3 & \textbf{59.4} & 32.5 & 20.3 & & 61.2 & \textbf{79.8} & 65.2 & 52.8 \\
    ZSCP {\footnotesize choice} (ours) & $1$ & & \textbf{39.8} & 41.4 & \textbf{37.7} & \textbf{23.7} & & 62.6 & 80.2 & \textbf{68.7} & 55.8 \\
    ZSCP {\footnotesize score} (ours) & $1$ & & 32.7 & 19.6 & 19.7 & 15.5 & & \textbf{84.9} & 60.7 & 67.9 & \textbf{63.0}  \\
  \bottomrule
  \end{tabular}
\end{table*}

%% file: tables/zero_shot_detection.tex
\begin{table}[t]
  \footnotesize
  \caption{Zero-shot object detection results as AP$_{50}$.}
  \label{tab:zeroshot_detection}
  \centering
    \setlength{\tabcolsep}{4pt}
  \begin{tabular}{@{}lcccc@{}}
    \toprule
    Class proposal & & ArtDL 2.0 \cite{milani2022proposals} & & IconArt \cite{gonthier2018weakly} \\
    \midrule
    CLIP & & 13.3 & & 6.8 \\
    InstructBLIP  & & 18.6 & & 7.9 \\
    NADA {\footnotesize(with ZSCP)} & & \textbf{21.8} & & \textbf{15.1} \\    
    \bottomrule
  \end{tabular}
\end{table}

%% file: tables/prompt_construction.tex
\begin{table*}[t]
    \footnotesize
    \centering
    \caption{AP$_{50}$ for different prompt construction methods across NADA systems with different class proposers.}
    \begin{tabular}{l c cccc c cccc}
        \toprule
         & & \multicolumn{4}{c}{ArtDL~2.0} & & \multicolumn{4}{c}{IconArt} \\
        \cline{3-6}
        \cline{8-11}
        Prompt & & ZSCP {\footnotesize choice} & ZSCP {\footnotesize score} & WSCP & Oracle & & ZSCP {\footnotesize choice} & ZSCP {\footnotesize score} & WSCP & Oracle \\
        \midrule
        Template & & 21.8 & 13.8 & 45.8 & 61.3 & & 7.8 & 12.1 & 11.7 & 15.2 \\
        Caption & & 20.2 & 12.6 & 42.5 & 58.0 & & 9.9 & 15.1 & 13.8 & 18.7 \\
        \bottomrule
    \end{tabular}
    \label{tab:prompt_construction}
\end{table*}

%% file: tables/prompt_ablation.tex
\begin{table}[t]
    \footnotesize
    \centering
    \caption{AP$_{50}$ for NADA {\footnotesize (with ZSCP)} on images with a \textit{single} class and images with \textit{multiple} classes.}
    \begin{tabular}{l c cc c cc}
        \toprule
         & & \multicolumn{2}{c}{ArtDL~2.0} & & \multicolumn{2}{c}{IconArt} \\
        \cline{3-4}
        \cline{6-7}
        Prompt & & single & multiple & & single & multiple \\
        \midrule
        Template & & 22.4 & 0.03 & & 5.2 & 8.1 \\
        Caption & & 20.7 & 0.02 & & 5.6 & 11.0 \\
        \bottomrule
    \end{tabular}
    \label{tab:prompt_ablation}
\end{table}

%% file: figures/attention_box_samples.tex
\begin{figure*}
    \centering

    \includegraphics[height=1.8cm, valign=c]{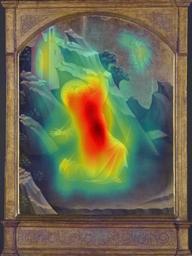}
    \includegraphics[height=1.8cm, valign=c]{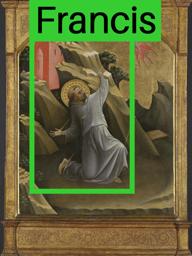}
    \hspace{7pt}
    \includegraphics[height=1.8cm, valign=c]{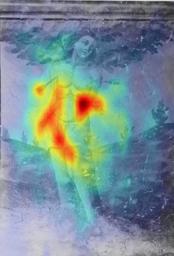}
    \includegraphics[height=1.8cm, valign=c]{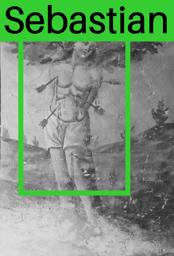}  
    \hspace{7pt}
    \includegraphics[height=1.8cm, valign=c]{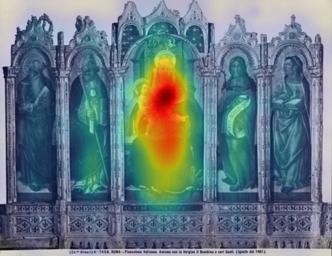}
    \includegraphics[height=1.8cm, valign=c]{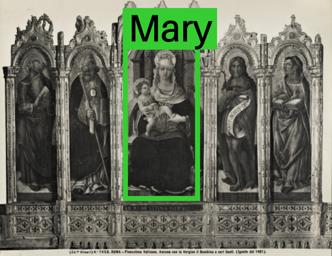}  
    \hspace{7pt}
    \includegraphics[height=1.8cm, valign=c]{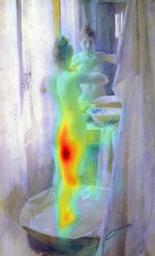}
    \includegraphics[height=1.8cm, valign=c]{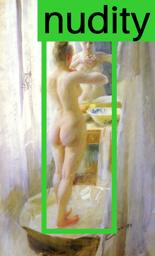}
    
    \vspace{2pt}
    \includegraphics[height=1.8cm, valign=c]{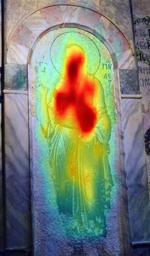}
    \includegraphics[height=1.8cm, valign=c]{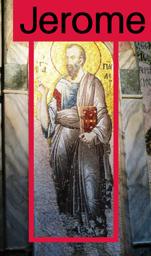}
    \hspace{7pt}
    \includegraphics[height=1.8cm, valign=c]{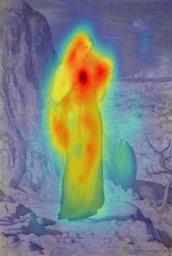}
    \includegraphics[height=1.8cm, valign=c]{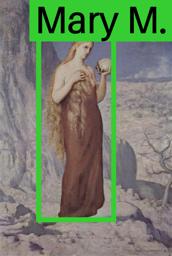}
    \hspace{7pt}
    \includegraphics[height=1.8cm, valign=c]{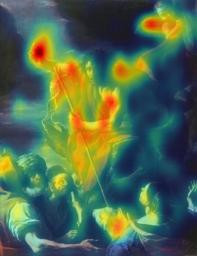}
    \includegraphics[height=1.8cm, valign=c]{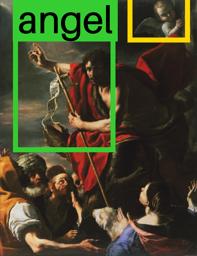}
    \hspace{7pt}
    \includegraphics[height=1.8cm, valign=c]{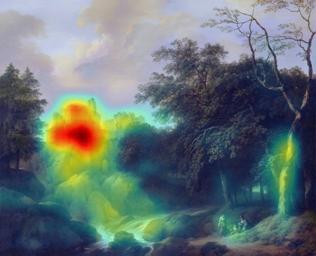}
    \includegraphics[height=1.8cm, valign=c]{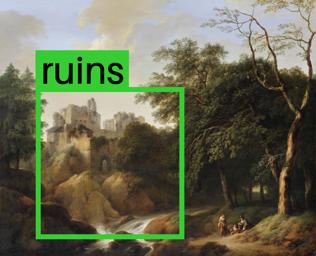}
    
    \caption{ArtDL~2.0 and IconArt test images overlaid with NADA {\footnotesize (with ZSCP)} attention maps and bounding boxes, shown in pairs. Redder areas indicate higher attention while bluer areas indicate lower attention. Correct model predictions are in green, incorrect model predictions are in red, and ground truth boxes when the predicted box has $< 0.5$ IoU with the ground truth are in yellow.}
    \label{fig:attention_box_samples}
\end{figure*}

%% file: sections/05-conclusion.tex
\section{Conclusion}

We introduced NADA, a method that applies diffusion models' knowledge of art to reduce the amount of supervision needed for object detection in paintings, specifically for weakly-supervised and zero-shot detection. Weakly-supervised NADA {\footnotesize (with WSCP)} competes closely with and outperforms other methods on art object detection, while zero-shot NADA {\footnotesize (with ZSCP)} is one of the first methods for zero-shot object detection in the domain of paintings. Detection performance improves when NADA's class proposer is always correct, demonstrating the importance of the class proposer to the whole pipeline. Prompting methods have varying effects on object detection depending on the target dataset. We use NADA's class-conditioned detector to detect objects in WikiArt images in the wild, demonstrating the detector's capacity to localize objects that are more commonly found in art images.

\vspace{-12pt}
\paragraph{Acknowledgements} This work is partly supported by JSPS KAKENHI No. JP23H00497 and JP22K12091, JST CREST Grant No. JPMJCR20D3, and JST FOREST Grant No. JPMJFR216O.

%% file: sections/06-appendix.tex
\appendix

\section*{Appendix}

\section{Datasets}

An overview of the two art object dection datasets, ArtDL~2.0  \cite{milani2022proposals} and IconArt \cite{gonthier2018weakly}, is provided in \cref{tab:datasets}. Both of the datasets consist of images of paintings containing Christian icons.

\section{WSCP training hyperparameters}

We present the hyperparameters for training the lightweight MLP in the WSCP in \cref{tab:hyperparameters}.

\section{Prompts}

We detail the various prompts used in NADA.

\subsection{ZSCP}
We present the prompts (choice and score) used to prompt the VLM to classify images in the ZSCP in \cref{tab:prompts_zscp}.

\subsection{Prompt construction}

We present the classes, prompts, templates used in the prompt construction for image reconstruction in the class-conditioned detector.

\paragraph{Class names} For each class in ArtDL, we use the title of its equivalent Wikipedia article, resulting in the following classes:

\begin{quote}
    \textit{Anthony of Padua}; \textit{John the Baptist}; \textit{Paul the Apostle}; \textit{Francis of Assisi}; \textit{Mary Magdalene}; \textit{Saint Jerome}; \textit{Saint Dominic}; \textit{Mary, mother of Jesus}; \textit{Saint Peter}; \textit{Saint Sebastian}
\end{quote}

Meanwhile for IconArt, we use the following texts for the classes:

\begin{quote}
    \textit{person} (equivalent to \textit{Saint Sebastian}), \textit{crucifixion of jesus}, \textit{angel}, \textit{mary}, \textit{baby} (equivalent to \textit{child jesus}), \textit{naked person} (equivalent to \textit{nudity}), \textit{ruins}
\end{quote}

\paragraph{Template} By default we insert the class in the simple prompt \texttt{A painting of [CLASS]}, where \texttt{[CLASS]} is the class being detected. For classes \emph{person}, \emph{baby}, and \emph{naked person}, we use \texttt{A painting of a [CLASS]}.

\paragraph{Caption} We prompt the same VLM used to classify the images in NADA {(\footnotesize with ZSCP)} to instead caption the images using the prompt \texttt{Describe the visual elements in the image in one sentence. Include the term "[CLASS]".} If the class is not found in the caption or is located at a part of the caption that is beyond the maximum input length of the diffusion model, we prepend the caption with the prompt \texttt{A painting of [CLASS].} formatted with the class name.

\section{Per-class detection results}

We present the AP$_{50}$ per class for ArtDL~2.0 in \cref{tab:artdl_full}. No class is detected the easiest or hardest across all experimental settings. When comparing methods, NADA {\footnotesize (with WSCP)} provides near consistent gains in AP$_{50}$ over NADA {\footnotesize (with ZSCP)}, improving AP$_{50}$ in all classes except for Mary and boosting detection performance within the same class by 24.1 AP$_{50}$ on average. Intuitively, Oracle has the best performance across all classes.

Per-class IconArt results are provided in \cref{tab:iconart_full}. NADA consistently detects Crucifixion of Jesus the best, but struggles to detect nudity and angel relative to other clases in all experimental settings. Furthermore, NADA {\footnotesize (with ZSCP)} outperforms NADA {\footnotesize (with WSCP)} on only four of the seven classes, with both methods having the same AP$_{50}$ on angel. Differences between class proposer are smaller, as NADA {\footnotesize (with ZSCP)} provides only a 1.2 AP$_{50}$ improvement over NADA {\footnotesize (with WSCP)}. While Oracle proves the best overall AP$_{50}$, it actually underperforms NADA on Crucufixion of Jesus, angel, and Mary.

\section{Qualitative analysis}

In \cref{fig:attention_box_samples} of the main paper, from left to right, top to bottom: samples 1, 2, 5, and 6 are from ArtDL~2.0 and samples 3, 4, 7, and 8 are from IconArt.

\input{tables/supplementary_tables}

%% file: tables/supplementary_tables.tex
\begin{table*}[tb]
  \caption{Details of the evaluation datasets. ArtDL 2.0 and IconArt provide different splits for classification and detection evaluation.}
  \label{tab:datasets}
  \setlength{\tabcolsep}{7pt}
  \centering
  \begin{tabular}{@{}lrr@{}}
    \toprule
     & ArtDL 2.0 \cite{milani2022proposals} & IconArt \cite{gonthier2018weakly} \\
    \midrule
    Type of art & Paintings & Paintings \\
    Type of objects & Christian icons & Christian icons\\
    Num. object classes & $10$ & $7$ \\
    Num. train images - classification & $21,673$ & $1,421$ \\
    Num. test images - classification & $2,632$ & $2,031$ \\
    Num. test images - detection & $808$ & $1,480$  \\
    Num. validation images - classification & $2,628$ & $610$ \\
    Num. validation images - detection & $1,625$ & -  \\
  \bottomrule
  \end{tabular}
\end{table*}

\begin{table*}[tb]
  \caption{Hyperparameters for training the MLP classifier in NADA {\footnotesize (with WSCP)}. LR is learning rate and WD is weight decay. }
  \setlength{\tabcolsep}{5pt}
  \label{tab:hyperparameters}
  \centering
  \begin{tabular}{@{}lrllrrrc@{}}
    \toprule
    Dataset & Layers & Classification & Loss & LR & WD & Classes\\
    \midrule
    ArtDL~2.0 \cite{milani2022proposals} & 2 & single-label & cross-entropy & $1\mathrm{e}{-4}$ & $0$ & 10 \\
    IconArt \cite{gonthier2018weakly} & 3 & multi-label & binary cross-entropy & $1\mathrm{e}{-3}$ & $1\mathrm{e}{-3}$ & 7 \\
  \bottomrule
  \end{tabular}
\end{table*}

\begin{table*}[tb]
    \centering
    \caption{Prompts used in the ZSCP of NADA {\footnotesize (with ZSCP)}. \texttt{[CLASSES]} refers to the list of classes.}
    \label{tab:prompts_zscp}
    \begin{tabular}{llp{8cm}}
        \toprule
        Prompt & Dataset & Contents \\
        \midrule
        Choice & ArtDL~2.0\cite{milani2022proposals} & \texttt{Who is in the painting? Choose from the following: [CLASSES]} \\
        \midrule
        Choice & IconArt \cite{gonthier2018weakly} & \texttt{Which of the options are in the painting? Choose from the following: [CLASSES]} \\
        \midrule
        Score & all datasets &  \texttt{Which of the Christian iconographic symbols are in the painting? Choose from the following: [CLASSES] 
        For each symbol, give a score from 0 to 1 of how confident you are.
        Put your answer in a dictionary first and then reason your answer. 
        Be as accurate as possible.
        If none of the symbols are present, output 'None'} \\
        \bottomrule
    \end{tabular}
\end{table*}

\begin{table*}[tb]
    \scriptsize
    \centering
    \caption{AP$_{50}$ for each class in ArtDL~2.0. \textit{Mean} refers to the overall AP$_{50}$ reported in the main paper.}
    \setlength{\tabcolsep}{4pt}
    \begin{tabular}{l ccccccccccc}
        \toprule
        Class Proposal & Antony of Padua & John the Baptist & Paul & Francis & Mary Magdalene & Jerome & Dominic & Mary & Peter & Sebastian & Mean \\
        \midrule
        NADA {\footnotesize (with WSCP)} & 29.5 & 35.1 & 26.7 & 50.7 & 60.1 & 58.3 & 51.3 & 55.5 & 40.2 & 51.5 & 45.8 \\
        NADA {\footnotesize (with ZSCP)} & 7.6 & 21.1 & 2.5 & 15.6 & 24.3 & 30.2 & 7.7 & 60.0 & 3.9 & 45.5 & 21.8 \\
        Oracle & 42.0 & 40.8 & 79.3 & 56.2 & 80.8 & 68.3 & 55.8 & 68.5 & 54.5 & 66.5 & 61.3 \\
        \bottomrule
    \end{tabular}
    \label{tab:artdl_full}
\end{table*}

\begin{table*}[tb]
    \scriptsize
    \centering
    \caption{AP$_{50}$ for each class in IconArt~2.0.}
    \setlength{\tabcolsep}{4pt}
    \begin{tabular}{l cccccccc}
        \toprule
        Class Proposal & Saint Sebastian & Crucifixion of Jesus & Angel & Mary & Child Jesus & Nudity & Ruins & Mean \\
        \midrule
        NADA {\footnotesize (with WSCP)} & 6.8 & 47.9 & 0.4 & 15.2 & 14.0 & 3.4 & 9.1 & 13.8 \\
        NADA {\footnotesize (with ZSCP)} & 11.7 & 43.1 & 0.4 & 20.7 & 15.0 & 2.2 & 12.3 & 15.1 \\
        Oracle & 21.0 & 45.8 & 0.3 & 20.3 & 17.5 & 5.4 & 20.3 & 18.7 \\
        \bottomrule
    \end{tabular}
    \label{tab:iconart_full}
\end{table*}